\documentclass{article}

\usepackage{microtype}
\usepackage{graphicx}
\usepackage{subfigure}
\usepackage{booktabs} 

\usepackage{hyperref}

\usepackage[accepted]{icml2025}

\usepackage{amsmath}
\usepackage{amssymb}
\usepackage{mathtools}
\usepackage{amsthm}
\usepackage[acronym,nogroupskip,nonumberlist,nopostdot]{glossaries}
\usepackage[capitalize,noabbrev]{cleveref}

\usepackage{multirow}
\usepackage{pgfplots}
\usepackage{pgfplotstable}
\pgfplotsset{compat=1.17}
\usepgfplotslibrary{statistics}
\usepackage{acronym}
\usepackage{float}

\newcommand{\myparagraph}[1]{\noindent \textbf{#1}}

\theoremstyle{plain}

\theoremstyle{definition}

\theoremstyle{remark}

\usepackage[disable,textsize=tiny]{todonotes}

\icmltitlerunning{RGBX-DiffusionDet: A Framework for Multi-Modal RGB-X Object Detection Using DiffusionDet}

\definecolor{mygreen}{RGB}{83,161,81}
\definecolor{negcolor}{RGB}{103,81,165}
\newcommand{\posacc}[1]{{\scriptsize\selectfont \color{mygreen}~(+#1)}}
\newcommand{\negacc}[1]{{\scriptsize\selectfont \color{negcolor}~(-#1)}}

\setcitestyle{numbers,square,comma}

\begin{document}

\twocolumn[
\icmltitle{RGBX-DiffusionDet: A Framework for Multi-Modal RGB-X Object Detection Using DiffusionDet}



\icmlsetsymbol{equal}{*}

\begin{icmlauthorlist}
\icmlauthor{Eliraz Orfaig}{equal,yyy}
\icmlauthor{Inna Stainvas}{equal,comp}
\icmlauthor{Igal Bilik}{yyy}
\end{icmlauthorlist}

\icmlaffiliation{yyy}{School of Electrical and Computer Engineering, Ben Gurion University of the Negev, Beer Sheva, Israel}
\icmlaffiliation{comp}{GE HealthCare, Haifa, Israel}

\icmlcorrespondingauthor{Eliraz Orfaig}{elirazo@post.bgu.ac.il}
\icmlcorrespondingauthor{Inna Stainvas}{Inna.Stainvas@gehealthcare.com}
\icmlcorrespondingauthor{Igal Bilik}{Bilik@bgu.ac.il}
\vskip 0.2in

\icmlkeywords{Multi-Modal Object Detection, Feature Fusion, Attention Mechanism, Diffusion Model}

]


\printAffiliationsAndNotice{}  

\begin{abstract}
This work introduces RGBX-DiffusionDet, an object detection framework extending the DiffusionDet model to fuse the heterogeneous 2D data (X) with RGB imagery via an adaptive multimodal encoder. To enable cross-modal interaction, we design the dynamic channel reduction within a convolutional block attention module (DCR-CBAM), which facilitates cross-talk between subnetworks by dynamically highlighting salient channel features. Furthermore, the dynamic multi-level aggregation block (DMLAB) is proposed to refine spatial feature representations through adaptive multiscale fusion. Finally, novel regularization losses that enforce channel saliency and spatial selectivity are introduced, leading to compact and discriminative feature embeddings. Extensive experiments using RGB-Depth (KITTI), a novel annotated RGB-Polarimetric dataset, and RGB-Infrared (M$^3$FD) benchmark dataset were conducted. We demonstrate consistent superiority of the proposed approach over the baseline RGB-only DiffusionDet. The modular architecture maintains the original decoding complexity, ensuring efficiency. These results establish the proposed RGBX-DiffusionDet as a flexible multimodal object detection approach, providing new insights into integrating diverse 2D sensing modalities into diffusion-based detection pipelines.
\end{abstract}

\section{Introduction} \label{sec:intro}
Vision-based object detection is a fundamental task in a wide range of modern applications, such as autonomous vehicles, robotic navigation, and medical imaging. These applications require high detection performance while maintaining low computational complexity. Object detection models can be generally classified into one-stage~\cite{yolov8_ultralytics,liu2016ssd} and two-stage~\cite{FasterRCNN} approaches, with the former enabling faster inference at the cost of degraded accuracy. Recently, a novel two-stage DiffusionDet model~\cite{chen2023diffusiondet} was introduced for object detection. It is initiated with random bounding box (BBox) proposals that are iteratively refined through a diffusion process. DiffusionDet has demonstrated superior detection performance compared to other two-stage frameworks.

The baseline DiffusionDet model was developed for object detection using RGB data only. This work introduces the RGBX-DiffusionDet, which incorporates auxiliary heterogeneous 2D data, referred to as 'X', to extend its detection capabilities. Since DiffusionDet is a diffusion-based model, the decoder's computational complexity increases with the number of refinement iterations. Therefore, it is essential to maintain the same decoder capacity as the RGB-only baseline. The proposed approach modifies only the model's encoder, preserving the decoder capacity intact despite fusing multi-modal features.

\begin{figure*}[t]
\begin{center}
\centerline{\includegraphics[width=0.75\textwidth]{ 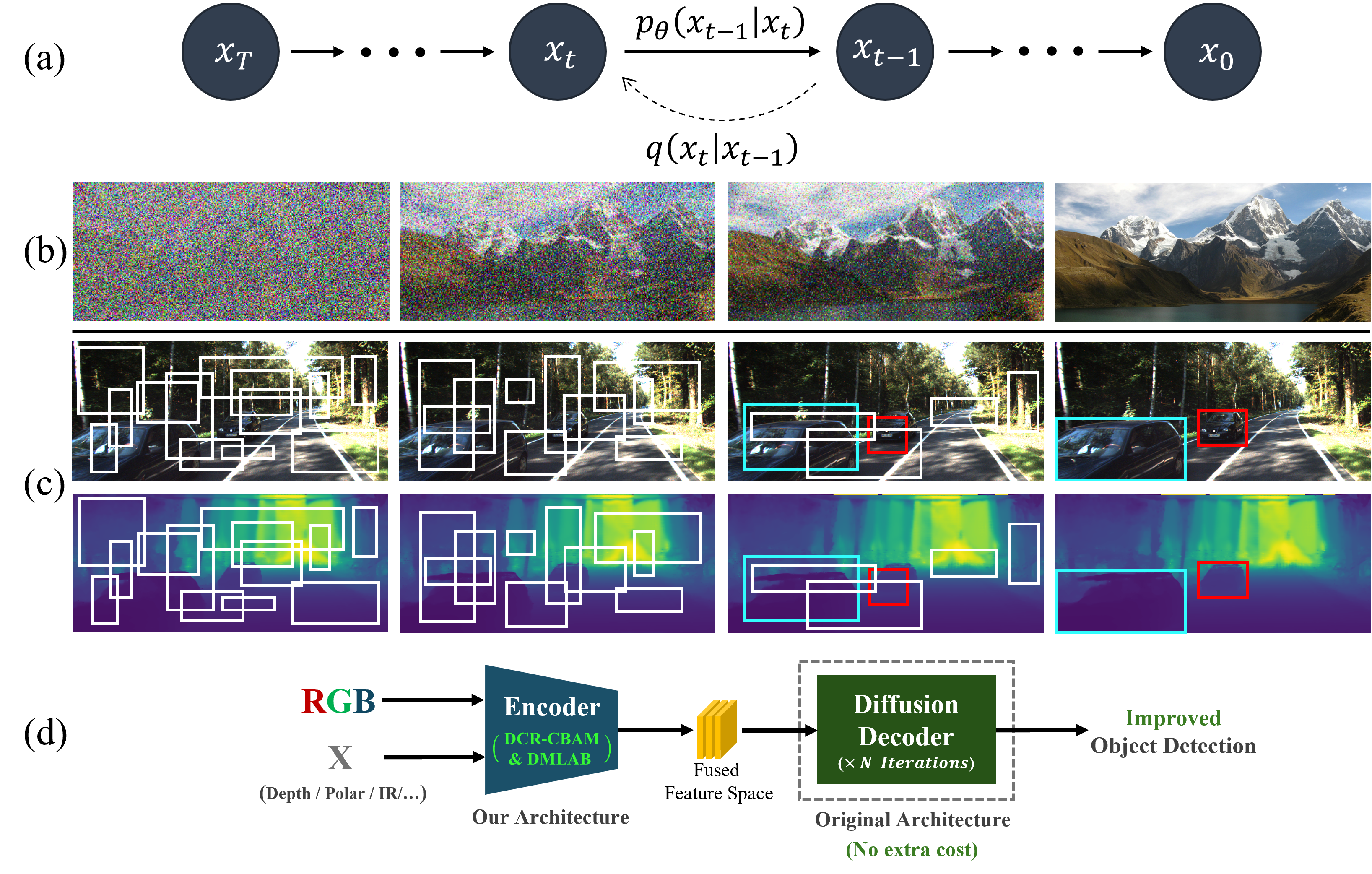}}
\vskip -0.15in
\caption{A conceptual representation of RGB-D diffusion model for object detection. (a) Schematic representation of diffusion model where $q$ is the diffusion process and $p_\theta$ is the learnable reverse process. (b) Exemplary diffusion process for image generation task. (c) Exemplary object detection process as a denoising diffusion process from noisy boxes to object boxes in the RGB-D dataset. (d) Overview of the proposed RGBX-DiffusionDet approach using modified encoder and the original DiffusionDet decoder for improved object detection.}
\label{fig1}
\label{icml-historical}
\end{center}
\vskip -0.3in
\end{figure*}

The proposed framework incorporates two new architectural components: 1) the dynamic channel reduction within a convolutional block attention module (DCR-CBAM) and 2) the dynamic multi-level aggregation block (DMLAB). The DCR-CBAM builds upon the widely used CBAM~\cite{woo2018cbam}, which applies lightweight channel and spatial attention operations. Our extension introduces a novel dynamic channel combination mechanism that leverages internal channel attention information to fuse RGB features seamlessly with auxiliary X features. The DMLAB addresses a key limitation of the baseline DiffusionDet decoder, where poorly initialized random BBoxes are challenging to refine through the diffusion process. By exploiting the spatially-aware feature refinements produced by DCR-CBAM, DMLAB dynamically aggregates multi-level features. Jointly, these components enable the model to capture richer contextual information and generate more discriminative feature representations, leading to significantly improved detection performance.

We evaluate the performance of the proposed approach on automotive object detection tasks, where the fusion of multiple sensor modalities is essential for achieving accurate and reliable detection. The detection performance superiority of the proposed RGBX-DiffusionDet was demonstrated using RGB-Depth (RGB-D)~\cite{Geiger2012CVPR}, RGB-Polarimetric (RGB-P)~\cite{Baltaxe_2023_BMVC}, and RGB-Infrared (RGB-IR)~\cite{Liu_2022_CVPR} datasets. This wide variety of sensing modalities demonstrates the generalization capabilities of the proposed approach. 

The key contributions of this work are:
\begin{itemize}
\item Development of RGBX-DiffusionDet, a modular and extensible framework that demonstrates the feasibility of integrating auxiliary 2D data into DiffusionDet.
\item Introduction of DCR-CBAM, a dynamic feature fusion approach.
\item Introduction of DMLAB, a dynamic feature aggregation operation, designed to enhance the performance of the diffusion decoding process.
\item Novel regularization losses that enforce channel saliency and spatial selectivity, enabling compact and discriminative feature embeddings.
\item The first use of pixel-aligned RGB-P data for object detection, including the generation of Bbox annotations, to motivate future research in multi-modal data processing.
\end{itemize}

This work takes a step toward bridging multi-modal sensing and diffusion-based object detection, contributing to the development of more adaptable and accurate vision models.

\section{Related Work}\label{RW}
\subsection{Multi-modal Fusion for Object Detection}
Integrating auxiliary sensor modalities, such as depth (RGB-D) or other types of 2D data (RGB-X), into object detection models has been studied in the literature~\cite{huang2022multi}. Recently introduced, DeepFusion~\cite{li2022deepfusion} enhances object detection by integrating deep LiDAR and camera features. The SparseFusion~\cite{xie2023sparsefusion} was introduced to selectively fuse sparse multi-modal representations in a unified 3D space. The fusion of RGB-IR data was also studied in the literature~\cite{Liu_2022_CVPR}. PIAFusion~\cite{tang2022piafusion}, SwinFusion~\cite{ma2022swinfusion}, and CDDFuse~\cite{zhao2023cddfuse} introduced various approaches to fuse auxiliary thermal, infrared, and domain-transformed features. This work leverages the recently introduced DiffusionDet model and establishes the general framework of RGB and auxiliary 2D data fusion.

\subsection{Attention Mechanisms as Feature Fusion}
Attention mechanisms have been adopted for multiple applications due to their ability to highlight critical information. Recently, CBAM~\cite{woo2018cbam} was proposed for feature refinement in convolutional neural networks (CNNs). This work proposes adopting it for feature fusion by applying spatial and channel-wise attention to emphasize the most informative features across multiple data sources. Similarly, methods such as SA-Gate~\cite{chen2020bi} and CMX~\cite{zhang2023cmx} have demonstrated the efficacy of attention mechanisms in fusing various types of feature maps. This work introduces the new CBAM-inspired, middle-level fusion approach, denoted as DCR-CBAM. 

\subsection{Polarimetric Data for Object Detection} 
Recently, the polarimetric data has been used for various perception tasks, such as free space detection and depth estimation~\cite{Baltaxe_2023_BMVC}. However, its utilization in object detection tasks remains limited~\cite{blin2020new}. Although polarimetric imaging provides unique information about surface textures and reflections not captured by RGB or depth data, it is a relatively unexplored modality for object detection. To the best of our knowledge, this work is the first to incorporate pixel-aligned RGB-P data into a detection framework, demonstrating this sensing modality's contribution to object detection performance, especially in challenging environments.

\subsection{DiffusionDet Extension Models} 
Recent efforts have extended DiffusionDet to various domains and challenges beyond its original formulation. Specifically, OrientedDiffDet~\cite{wang2024orienteddiffdet} adapts the framework to oriented object detection in aerial imagery, while LSKDiffDet~\cite{sharshar2023lskdiffdet} combines DiffusionDet with the LSKNet backbone to enhance performance on densely packed objects. DifAda~\cite{difada2024adaptive} introduces adaptive denoising techniques targeted at underwater scenarios, demonstrating environmental robustness. In addition, DetDiffusion~\cite{wang2024detdiffusion} explores diffusion models for joint data generation and perception, focusing on improving generative controllability. Other studies focus on optimization strategies within the DiffusionDet process itself, such as the use of deformable sigmoid variance and adjustable sampling~\cite{deformable2025sigmoid}. Parameter-efficient adaptations have also been explored through low-rank fine-tuning in cross-domain few-shot settings~\cite{talaoubrid2025lora}. These diverse approaches underscore the flexibility of the diffusion-based object detection paradigm. In contrast to these domain-specific or architectural extensions, the proposed RGBX-DiffusionDet targets multi-modal RGB-X object detection, introducing cross-modal fusion mechanisms to enhance perception across heterogeneous visual domains.

\section{The Proposed Approach}\label{Sec:PA}
\begin{figure*}[t]
\begin{center}
\centerline{\includegraphics[width=\textwidth]{ 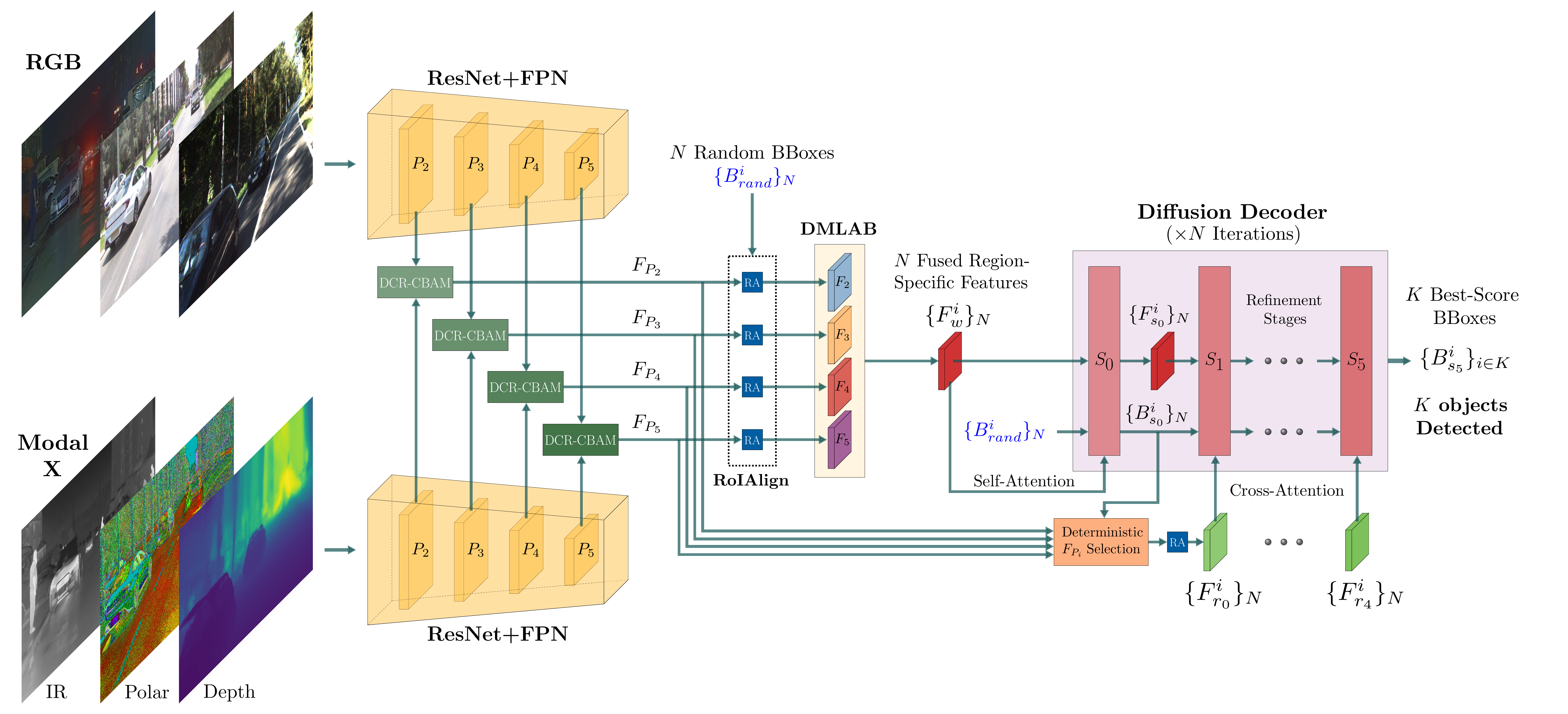}}
\caption{Pipeline overview of RGBX-DiffusionDet. Separate ResNet+FPN encoders process RGB and X modalities. Feature maps are fused using DCR-CBAM and aggregated with DMLAB before being refined by the original DiffusionDet decoder}
\label{arch_fig}
\end{center}
\vskip -0.2in
\end{figure*}

\subsection{Background}
This subsection provides the background for the concepts used in this work.

\subsubsection{Diffusion Model for Object Detection} 
Diffusion models are inspired by nonequilibrium thermodynamics principles~\cite{sohl2015deep} and form a Markovian chain through a forward diffusion process that systematically introduces noise into BBox, $x_t$, at time $t$:
\begin{equation}
q(x_t|x_0) = \mathcal{N}(x_t | \sqrt{\bar{\alpha}_t} x_0, (1 - \bar{\alpha}_t) \mathbf{I})\;,\label{Eq1}
\end{equation}
where $q(x_t | x_0)$ is the forward diffusion process, modeled as a Gaussian distribution, $\mathcal{N}(\cdot)$, with mean vector, $\sqrt{\bar{\alpha}_t} x_0$, and the covariance matrix, $(1 - \bar{\alpha}_t) \mathbf{I}$, where $x_0$ is the initial BBox, $\bar{\alpha}_t$ is the cumulative noise scaling factor, and $\mathbf{I}$ is the identity matrix ensuring isotropic noise. This process incrementally shifts and reshapes the BBoxes, transitioning them from precise ground truth annotations to progressively noisier representations.

During training, a neural network, $f_{\theta}(x_t, t | z_x)$, parameterized by $\theta$, learns to reverse this diffusion process, predicting the original BBox, $x_0$, from $x_t$ given the extracted features, $z_x$. The training objective is to minimize the loss:
\begin{equation}
\mathcal{L}_{\text{train}} = \frac{1}{2} \|f_{\theta}(x_t, t | z_x) - x_0\|^2\;,
\end{equation}
which guides the network in minimizing the difference between the reconstructed and ground truth BBoxes.

\subsubsection{Convolutional Block Attention Module}
The CBAM~\cite{woo2018cbam} enhances intermediate features representation of CNNs by using attention mechanisms in both channels and spatial dimensions. Given an input feature map, $\mathbf{F} \in \mathbb{R}^{C \times H \times W}$, CBAM refines it via the following two stages:
\begin{align}
\mathbf{F}'  &= \mathbf{M}_c(\mathbf{F}) \odot \mathbf{F} \label{Eq3}\;, \\
\mathbf{F}'' &= \mathbf{M}_s(\mathbf{F}') \odot \mathbf{F}'\;, \label{Eq4}
\end{align}
where, $\odot$ denotes element-wise multiplication, $\mathbf{M}_c \in \mathbb{R}^{C \times 1 \times 1}$ is the \textit{channel attention map}, and $\mathbf{M}_s \in \mathbb{R}^{1 \times H \times W}$ is the \textit{spatial attention map}, defined as:
\begin{align}
\mathbf{M}_c(\mathbf{F}) &= \sigma \bigl(MLP(\mathbf{F^c}_{avg})+ MLP(\mathbf{F^c}_{max}) \bigr)\;, \label{Eq5} \\
\mathbf{M}_s(\mathbf{F}) &= \sigma\bigl(conv_{k \times k}([\mathbf{F^s}_{avg};\mathbf{F^s}_{max}])\bigr)\;, \label{Eq6}
\end{align}
were $\sigma(\cdot)$ is the activation function, and $\mathbf{F}^c_{\text{avg}},\mathbf{F}^c_{\text{max}} \in \mathbb{R}^{C \times 1 \times 1}$ are the spatially average-pooled and max-pooled features, capturing global channel-wise information. These features are processed through a shared multi-layer perceptron (MLP) to produce the channel attention map. The channel-wise pooled features, $\mathbf{F}^s_{\text{avg}},\mathbf{F}^s_{\text{max}} \in \mathbb{R}^{1 \times H \times C}$, are concatenated and passed through a convolutional layer to generate the spatial attention map.
The channel attention mechanism focuses on "what" features are important, and the spatial attention mechanism identifies "where" they are significant. Despite its simplicity, CBAM provides significant performance improvements with minimal computational overhead, making it attractive for various CNN architectures.

\subsection{The Proposed Architecture}
Diffusion models are computationally challenging due to their iterative nature. Extending the baseline DiffusionDet model to support RGB-X input could further increase the computational complexity during the iterative decoding process. This work addresses this challenge by introducing an architecture with a multimodal encoder that processes RGB-X input while maintaining the diffusion decoder unchanged. The proposed RGBX-DiffusionDet approach performs separate feature extraction for each modality, followed by a feature pyramid network (FPN) to capture multi-scale features across different layers (Figure~\ref{arch_fig}). This subsection first introduces the DCR-CBAM, designed to enable dynamic feature fusion for each matching pair of FPN levels, [$\mathbf{P}^{RGB}_i$; $\mathbf{P}^X_i$], optimizing the integration of diverse features. Next, the DMLAB is introduced to perform dynamic aggregation of these fused features to refine the overall feature representation.

\begin{figure*}[t]
\begin{center}
\centerline{\includegraphics[width=0.9\textwidth]{ 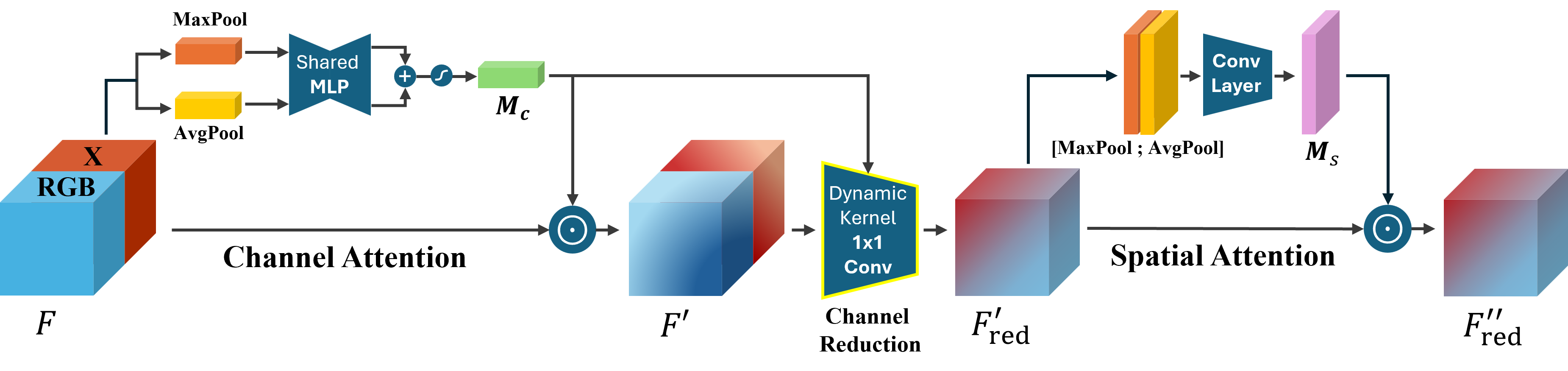}}
\caption{Schematic representation of the cross-modality feature refinement using dynamic channel reduction within CBAM for RGB-X fusion.}
\label{dcr-cbam-fig}
\end{center}
\vspace{-1.5em}
\end{figure*}

\subsubsection{Dynamic Channel Reduction CBAM} \label{sec:DCR-CBAM}
Given the concatenated feature maps from multiple modalities, $\mathbf{F} \in \mathbb{R}^{C \times H \times W}$, the channel attention map, $\mathbf{M}_{c}(\mathbf{F})$, in~\eqref{Eq5} is computed to produce a stage-one refined feature map, $\mathbf{F}'$  in~\eqref{Eq3}. However, this concatenated representation may introduce redundancy into the feature space, leading to an unnecessary increase in network complexity and further computational overhead throughout the subsequent stages.

The primary contribution of this work is leveraging the information encoded in the channel attention map, $\mathbf{M}_c$, to reduce the dimensionality of the combined high-dimensional feature map, $\mathbf{F}'$, from $C$ channels to a fixed $C'$ channels, where $C' < C$ and is a model hyper-parameter. This is accomplished using a data-dependent linear operator that is generated adaptively for each input. Specifically, the 1D attention vector $\mathbf{M}_c$ is transformed into a dynamic reduction matrix, $\mathbf{W}_{\text{red}}(\mathbf{M}_c) \in \mathbb{R}^{C' \times C}$. The refined feature map $\mathbf{F}'$ is then projected into a lower-dimensional space by applying this reduction matrix:
\begin{equation}
\mathbf{F}'_{\text{red}} = \mathbf{W}_{\text{red}} \mathbf{F}'\;, \label{Eq9}
\end{equation}
where, $\mathbf{F}'_{\text{red}} \in \mathbb{R}^{C' \times H \times W}$ is the reduced feature map. 
The $\mathbf{W}_{\text{red}}(\mathbf{M_c})$ is constructed  using a single fully connected layer to project the channel attention map, $\mathbf{M}_{c}$, into a $(C'C)$-dimensional space:
\begin{equation}
\mathbf{Z} = \sigma \left( \mathbf{W}_{\text{proj}}\mathbf{M}_{c} + \mathbf{b}_{\text{proj}} \right)\;, \label{Eq7}
\end{equation}
where $\mathbf{W}_{\text{proj}} \in \mathbb{R}^{(C'C) \times C}$ and $\mathbf{b}_{\text{proj}} \in \mathbb{R}^{C'C}$ are trainable parameters. The vector, $\mathbf{Z}$, is then reshaped into the desired $C' \times C$ matrix:
\begin{equation}
\mathbf{W}_{\text{red}} = \text{reshape}(\mathbf{Z}, C, C')\;, \label{Eq8}
\end{equation}
which acts as a dynamic linear operator compressing the channel dimensions of the original feature map. This process can be interpreted as a $1 \times 1$ convolution with dynamic kernels, where the weights are conditioned on the input attention vector (Figure \ref{dcr-cbam-fig}).

The reduced feature map, $\mathbf{F}'_{\text{red}}$, is then mapped into the spatial attention map, $\mathbf{M}_{s}(\mathbf{F}'_{\text{red}})$ in~\eqref{Eq6}, resulting in the refined feature map of stage two, $\mathbf{F}''_{\text{red}} \in \mathbb{R}^{C' \times H \times W}$, similar to~\eqref{Eq4}.
At this stage, $\mathbf{F}''_{\text{red}}$ appears as a dynamically fused representation, where information from multiple input modalities is integrated to enhance the expressiveness of the features while reducing redundancy.

\subsubsection{Dynamic Multi-Level Aggregation Block}
The objective of each step, $S_i$, within the decoder in Figure~\ref{arch_fig}, is to refine the object BBox proposals. These steps require $N$ proposal BBoxes and $N$ corresponding feature sets. For simplicity, $N=1$ is assumed in the following discussion. In each refinement step, $S_i$, cross-attention is applied between two sets of features: $F_{S_{i-1}}$, refined features obtained in the previous step, $S_{i-1}$ and $F_{r_{i-1}}$, raw FPN features of a specific level $r_{i-1}$. This level is selected by a deterministic function and denotes the FPN level selected according to the size of the BBox predicted in $S_{i-1}$ step, as shown in Figure~\ref{arch_fig}. This cross-attention operation refines the features associated with the previously predicted object area using updated features derived from the most recent BBox prediction, which are subsequently used by $S_i$. 

In the initial step, $S_0$, the proposal BBox is randomly generated, which means its sizes may not accurately reflect the true object scale. This randomness can lead to selecting an FPN level that does not effectively represent the objects in the scene. Although proposal boxes tend to converge in later stages, poor initializations may hinder early refinement and delay convergence.

\begin{figure*}[t]
\begin{center}
\centerline{\includegraphics[width=0.95\textwidth]{ 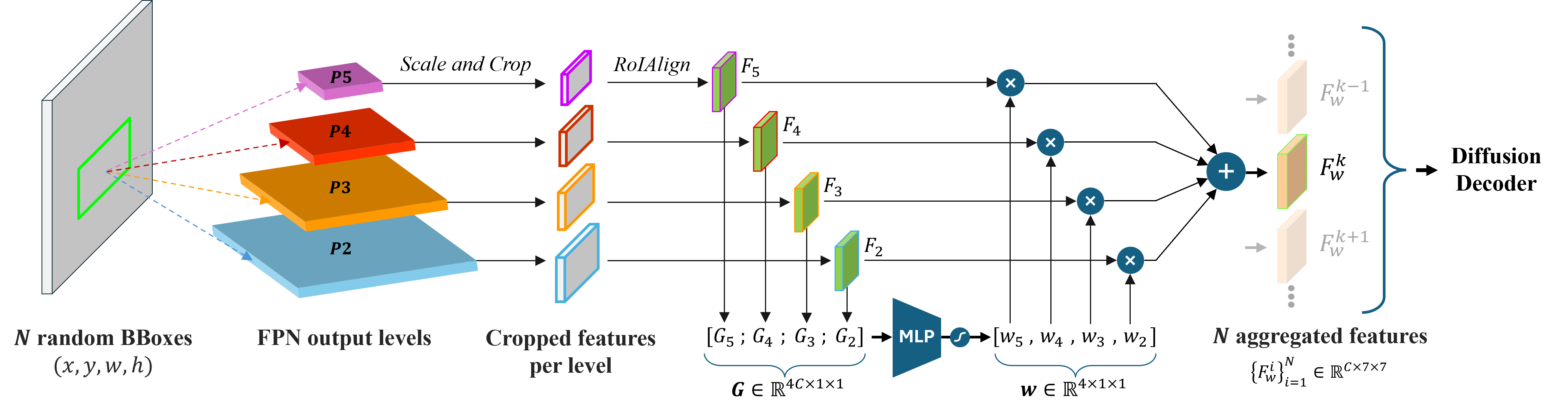}}
\caption{The DMLAB pipeline for a single proposal BBox. RoI-aligned features are extracted from each FPN level, weighted by MLP-derived scores, and fused into a unified representation.}
\label{dmlab_fig}
\end{center}
\vspace{-1.5em}
\end{figure*}

The proposed approach addresses this limitation by a multi-level aggregation strategy that dynamically incorporates features from all FPN levels instead of relying on a single level (Figure~\ref{dmlab_fig}). Let $\mathcal{I}$ be the set of indices corresponding to the FPN levels, $P_i$. Each pair of features, $\{P_i^{RGB},P_i^X\}_{i \in \mathcal{I}}$ is fused using DCR-CBAM, $F_{P_i}$. Given $N$ randomly initialized BBoxes, these fused features are then used to extract RoI-aligned crops at each scale, generating level-specific features per proposal. This step ensures that spatially relevant features are captured from all levels for every BBox. The RoIAlign~\cite{girshick2015fast} block further standardizes the spatial dimensions across scales, producing features $F_i \in \mathbb{R}^{C \times 7 \times 7}$.

A global descriptor, $G_i \in \mathbb{R}^{C \times 1 \times 1}$, for each feature map, is calculated using global average pooling to capture comprehensive feature information across all $\{F_i\}_{i \in \mathcal{I}}$:
\begin{equation}
G_i = \frac{1}{HW} \sum_{h=1}^{H} \sum_{w=1}^{W} F_i(:,h,w)\;.
\end{equation}
These descriptors are concatenated into a unified vector, $G \in \mathbb{R}^{(C |\mathcal{I}|) \times 1 \times 1}$, where $|\mathcal{I}|$ is the number of FPN levels. This unified vector is then processed by an MLP model to produce scalar weights, $\{w_i\}_{i \in \mathcal{I}}$, which are normalized using a softmax function. The resulting weights are then used to dynamically compute a fused feature representation, $F_w$, as a weighted sum of the features:
\begin{equation}
F_w = \sum_{i \in \mathcal{I}}w_iF_i\;.
\end{equation}
This adaptive multi-scale representation, $F_w$, integrates information from all FPN levels, ensuring a comprehensive and balanced feature set per the proposed BBox. In the initial step, the model uses $F_w$ to mitigate suboptimal initialization and as a result, improve BBoxes refinement across subsequent iterations. This process is performed independently for each $N$ proposal BBoxes, generating $N$ distinct feature representations, $\{F_w^i\}_{N}$. This ensures that each BBox benefits from an adaptive multi-scale representation.

\section{Training Loss}
The proposed total training loss is:
\begin{equation}
\mathcal{L}_{\text{total}} = \underbrace{\lambda_1\mathcal{L}_{\text{label}}}_{\substack{\text{Classifi-} \\ \text{cation}}} + \underbrace{\lambda_2\mathcal{L}_{\text{bbox}} + \lambda_3\mathcal{L}_{\text{giou}}}_{\text{Regression}} + \underbrace{\theta_{M_{c}} + \theta_{M_{s}}}_{\substack{\text{DCR-CBAM}\;, \\ \text{Regularization}}}
\end{equation}
where $\lambda_i, i=1,2,3$ control the balance between loss terms. These and other hyperparameters introduced in this chapter are detailed in Appendix A.1. The following subsections detail each loss component.

\subsection{Classification and Regression Losses}
The classification and regression losses are retained from the baseline DiffusionDet to ensure consistency with prior work. The classification term includes a focal loss, $\mathcal{L}_{\text{label}}$, which addresses class imbalance by prioritizing harder, misclassified examples~\cite{lin2017focal}. For BBox regression, it is proposed to optimize the precision and spatial alignment of predicted boxes by using the conventional, $L_1$, loss, $\mathcal{L}_{\text{bbox}}$, and the generalized IoU loss, $\mathcal{L}_{\text{giou}}$~\cite{rezatofighi2019generalized}. These losses efficiently balance classification and localization tasks and are widely adopted in modern object detection frameworks.

\subsection{DCR-CBAM Regularization Losses}\label{sec:regul_losses}
The functionality of the DCR-CBAM module is enhanced by introducing a) \emph{channel saliency regularization}, $\theta_{M_c}$, for efficient channel dimensionality reduction, and b) \emph{object-focused spatial regularization}, $\theta_{M_s}$, for the focus on target objects refinement. These regularization terms are applied for each of the $K$ DCR-CBAM blocks in Figure~\ref{arch_fig}.

\myparagraph{Channel saliency regularization} is defined as:
\begin{equation}
\theta_{M_c} = \sum_{k=1}^{K} \lambda_c^k  \| \mathbf{M}_c^k \|_1\;, \label{channel_loss}
\end{equation}
where the $L_1$-norm is applied to the channel attention maps, $\mathbf{M}_c^k$, weighted by a scaling factor, $\lambda_c^k$, to control each level's contribution. This regularization term encourages the network to suppress less informative channels while emphasizing the more important ones. As a result, it provides a sparser and more interpretable attention distribution. The efficiency of the channel reduction operation is enhanced by incorporating this constraint, ultimately leading to improved detection performance.

\myparagraph{Object-focused spatial regularization} guides spatial attention maps to focus on the object regions hinted at by annotated BBoxes. Specifically, for each input image, a binary mask, $B_{\text{mask}} \in \{0,1\}^{1 \times H \times W}$, is defined, where each pixel is set to $1$, if it lies within any of the ground-truth BBoxes, and $0$ otherwise. Since each instance of DCR-CBAM operates at a distinct spatial scale, the spatial attention map, $\mathbf{M}_s^k$, at scale $k$ has a resolution, $(H_k, W_k)$. $B_{\text{mask}}$ is bilinearly interpolated and quantized to produce binary $\tilde{B}_{\text{mask}}^k$ map, matching the resolution of $\mathbf{M}_s^k$ and ensuring mask alignment.

The following regularization term encourages the spatial attention map activation within the object regions and minimizes activation elsewhere: 
\begin{equation} 
\mathcal{M}_s^k = (\mathbf{M}_s^k \odot \tilde{B}_{\text{mask}}^k) - \tilde{B}_{\text{mask}}^k \label{eq:spat_at_reg}\;. \end{equation} 
Intuitively, this regularization guides the spatial attention to align with annotated object regions, encouraging early object localization within the attention mechanism.
The total spatial regularization loss is then computed as:
\begin{equation} 
\theta_{M_s} = \sum_{k=1}^{K} \lambda_s^k \| \mathcal{M}_s^k \|_2^2\;,
\end{equation}
where $ \lambda_s^k $ balances the contribution of each resolution level.
This implicit pre-detection phase refines feature representations prior to the main detection process and enhances localization accuracy. 

\section{Datasets}
The comprehensive evaluation of the proposed RGBX-DiffusionDet requires datasets with RGB imagery complemented by additional 2D sensor data (denoted as X), and BBox annotations for supervised object detection training. This work considers three automotive-related datasets that satisfy these criteria: the KITTI dataset for RGB-D experiments~\cite{Geiger2012CVPR}, a novel pixel-aligned RGB-P dataset~\cite{Baltaxe_2023_BMVC}, annotated and labeled in this work, and the M$^{3}$FD dataset for RGB-IR experiments~\cite{Liu_2022_CVPR}. Details on the datasets are provided in Appendix A. 

\begin{table*}[t]
\caption{Detection performance of the RGB-only DiffusionDet baseline compared with the proposed RGBX-DiffusionDet approach using the RGB-D, RGB-P, and RGB-IR data across three considered datasets. A dash "-" indicates object categories not present in a given dataset. The best results are highlighted in bold.}
\label{rgbxDetPerf}
\begin{center}
\begin{small}
\begin{sc}
\begin{tabular}{l|c|c@{\hskip 10pt}c@{\hskip 10pt}c@{\hskip 10pt}c@{\hskip 10pt}c@{\hskip 10pt}c@{\hskip 10pt}c@{\hskip 10pt}c@{\hskip 10pt}c@{\hskip 10pt}c@{\hskip 10pt}c}
\toprule
Method & AP$_{50:95}$ & \scriptsize{Pedestrian} & \scriptsize{Car} & \scriptsize{Truck} & \scriptsize{Cyclist} & \scriptsize{Van} & \scriptsize{Person Sitting} & \scriptsize{Tram} & \scriptsize{Misc} & \scriptsize{Bus} & \scriptsize{Lamp} \\
\midrule
RGB (baseline) & 67.0 & 45.1 & 78.8 & \textbf{89.1} & 66.0 & 78.9 & 36.8 & 70.6 & 70.8 & - & - \\
RGB-D (ours) & \textbf{69.2} & \textbf{46.9} & \textbf{80.3} & 88.8 & \textbf{70.9} & \textbf{79.0} & \textbf{42.0} & \textbf{73.1} & \textbf{71.8} & - & - \\
\midrule
RGB (baseline) & 52.8 & \textbf{44.5} & 76.6 & 45.5 & 46.8 & \textbf{50.9} & - & - & - & - & - \\
RGB-P (ours) & \textbf{54.2} & 42.4 &\textbf{77.6} & \textbf{54.4} & \textbf{48.8} & 47.7 & - & - & - & - & - \\
\midrule
RGB (baseline) & 54.1 & 39.9 & 63.2 & 59.3 & 45.4 & - & - & - & - & 67.9 & 49.0 \\
RGB-IR (ours) & \textbf{58.1} & \textbf{53.2} & \textbf{64.0} & \textbf{61.2} & \textbf{48.7} & - & - & - & - & \textbf{71.5} & \textbf{49.8} \\
\bottomrule
\end{tabular}
\end{sc}
\end{small}
\end{center}
\vskip -0.15in 
\end{table*}

\subsection{RGB-Depth Dataset}
The widely-used KITTI dataset~\cite{Geiger2012CVPR}, which was collected for autonomous driving research and contains RGB images along with LIDAR point cloud data and corresponding BBox annotation, is used in this work. In this dataset, the LIDAR data is represented as 3D point clouds, which are not directly compatible with the proposed 2D image-based architecture. Therefore, the LIDAR 3D point clouds are first projected onto the 2D image plane, and next, a depth completion algorithm~\cite{tang2024bilateral} was used to fill in the missing depth values.

\subsection{RGB-Polarimetric Dataset}
The generalization capabilities of the proposed approach are further evaluated using a novel pixel-aligned RGB-P dataset with annotated object BBoxes~\cite{Baltaxe_2023_BMVC}. This dataset contains both conventional RGB images and corresponding polarimetric data. The polarimetric data enhances the RGB images by providing additional information about surface properties and scene composition. To the best of our knowledge, there is no publicly available automotive RGB-P dataset with object detection annotations. Therefore, manual labeling was performed in this work to create ground-truth bounding boxes, which are publicly available\footnote{\url{https://github.com/ElirazO/RGB-Polarimetric_BBox_annotations.git}}. Appendix~\ref{PolarApnx} summarizes the theoretical background on polarimetric imaging and its properties.

\subsection{RGB-Infrared Dataset}
The M$^{3}$FD dataset~\cite{Liu_2022_CVPR}, which is a commonly used benchmark for multimodal feature object detection, is considered in this work. This dataset provides both RGB and infrared (IR) imagery, which is suitable for the evaluation of detection performance in low-light conditions. By incorporating IR data, M$^{3}$FD allows for rigorous testing of fusion-based approaches and serves as a reliable reference for comparison of the state-of-the-art multimodal detection approaches.

\vspace{1em}
Across the three datasets, the evaluation covers a wide range of sensor modalities, lighting conditions, and environments, from bright daytime scenes to challenging low-light imagery. This diversity demonstrates the robustness of RGBX-DiffusionDet in realistic and unconstrained detection scenarios.

\section{Experiments}\label{Exp}
This section evaluates the proposed RGBX-DiffusionDet framework. 
The experiments reported in this section were conducted using a single diffusion decoding iteration for both the baseline DiffusionDet and the proposed RGBX-DiffusionDet models. This configuration ensures a fair and consistent comparison across modalities and architectural variants. The impact of increasing the number of diffusion iterations is examined in Subsection \ref{subsec: dif_iters} and is not included in the main performance comparisons.
\subsection{RGB-X Detection Performance}

Table~\ref{rgbxDetPerf} summarizes the detection performance of the proposed RGBX-DiffusionDet approach using RGB-D, RGB-P, and RGB-IR datasets, compared with the baseline DiffusionDet approach using the RGB data only. Notice the superiority of the proposed RGBX-DiffusionDet framework, which consistently outperforms the RGB-only baseline model in all considered datasets. RGB-D data enables the improvement of smaller or partially occluded object detection performance. Notice this detection performance improvement for \emph{Pedestrian} ($+1.8$ AP) and \emph{Cyclist} ($+4.9$ AP) classes. RGB-P data contains information on object material, enabling the improvement of large and textured object detection performance. Notice this improvement for \emph{Truck} ($+8.9$ AP) and \emph{car} ($+1.0$ AP). RGB-IR data provides significant information in low-visibility scenarios. It enables significant detection performance improvements for heat-emitting objects like \emph{Pedestrian} ($+13.3$ AP). These results demonstrate the generalization capabilities of the proposed approach. 

\subsection{Comparison of Fusion-Based Detection Models}
The performance of the RGBX-DiffusionDet fusion approach is evaluated using the M$^{3}$FD benchmark dataset. Table~\ref{tab:M3FD} summarizes the results, including AP$_{50:95}$, which assesses detection performance for multiple IoU thresholds, and AP$_{50}$, which evaluates accuracy at an IoU of $0.5$. The RGBX-DiffusionDet approach achieved an AP$_{50:95}$ of $58.1$ and an AP$_{50}$ of $88.8$, outperforming the state-of-the-art fusion approaches. Notice that the proposed generic approach outperforms all the considered approaches, including those such as SwinFusion~\cite{ma2022swinfusion} and PIAFusion~\cite{tang2022piafusion}, which were specifically designed for RGB-IR data. Table~\ref{one_stage_comparison} compares the detection performance of the RGBX-DiffusionDet to one of the one-stage baseline models, evaluated using the RGBD dataset. The performance of the proposed approach is comparable with this baseline detector adjusted to a multi-modal RGB-D case.
Note that while one-stage models provide strong real-time performance, the iterative refinement phase of the diffusion process in two-stage models enables improvement of feature utilization and detection consistency.

\begin{table}[t]
\caption{Object detection performance of the proposed approach compared with state-of-the-art prior approaches, evaluated using the M$^{3}$FD dataset~\cite{zhange2e}. The best performance is marked in bold.}
\label{tab:M3FD}
\begin{center}
\begin{small}
\begin{sc}
\begin{tabular}{lcc}
\toprule
Method & AP$_{50:95}$ & AP$_{50}$ \\
\midrule
Tardal~\cite{Liu_2022_CVPR}           & 56.0          & 86.0           \\
SwinFusion~\cite{ma2022swinfusion}    & 56.1          & 85.8           \\
DIDFusion~\cite{zhao2020didfuse}      & 56.2          & 86.2           \\
CDDFuse~\cite{zhao2023cddfuse}        & 56.5          & 86.0           \\
MetaFusion~\cite{zhao2023metafusion}  & 56.8          & 86.7           \\
U2Fusion~\cite{xu2020u2fusion}        & 57.1          & 87.1           \\
PIAFusion~\cite{tang2022piafusion}    & 57.2          & 87.3           \\
ICAFusion~\cite{shen2024icafusion}    & 57.2          & 87.4           \\
RGBX-DiffusionDet (Ours)              & \textbf{58.1} & \textbf{88.8}  \\
\bottomrule
\end{tabular}
\end{sc}
\end{small}
\end{center}
\vskip -0.15in 
\end{table}

\begin{table}[t]
\caption{Detection performance of the proposed RGBX-DiffusionDet compared with the adjusted RGBX YOLOv8n, evaluated using the RGB-D dataset. The experiments were conducted with a batch size of $4$, where only the backbone was refined.
}
\label{one_stage_comparison}
\begin{center}
\begin{small}
\begin{sc}
\begin{tabular}{lccc}
\toprule
Approach & AP$_{50:95}$ & AP$_{50}$\\
\midrule
RGBD YOLOv8n (Adjusted) & 68.7 & 91.1\\
RGBX-DiffusionDet (Ours) & \textbf{69.2} & \textbf{91.3}\\
\bottomrule
\end{tabular}
\end{sc}
\end{small}
\end{center}
\vskip -0.15in 
\end{table}

\subsection{Inference Efficiency}\label{sec:inference_time}
The computational efficiency of the proposed RGBX-DiffusionDet is evaluated by comparing the inference times with the baseline RGB-only DiffusionDet model. The average inference time per image was measured for $748$ test images. Table~\ref{tab:inference_time} summarizes the results for two different numbers of diffusion steps $3$ and $10$. Despite the additional complexity of the encoder for fusing RGB and depth modalities, RGBX-DiffusionDet introduces only a slight increase in the processing time. As the number of diffusion iterations increases, the additional runtime decreases due to reducing the overhead proportionally to more extensive decoding.

Table~\ref{tab:inference_time} shows that RGBX-DiffusionDet maintains practical inference times even in multi-modal settings. Furthermore, as shown in Subsection \ref{subsec: dif_iters}, increasing the number of diffusion steps consistently improves detection performance, indicating that the resulting accuracy gains justify the additional inference time.

\begin{table}[t]
\begin{center}
\caption{Average inference time per image in RGB-D case. The additional runtime is reduced as the number of diffusion steps increases.}
\vspace{-0.5em}
\begin{small}
\begin{sc}
\resizebox{\columnwidth}{!}{%
\begin{tabular}{cccc}
\toprule
Steps & DiffusionDet & RGBX-DiffusionDet & Gap \\
\midrule
3  & 0.0548 s & 0.0681 s & 24.2\% \\
10 & 0.1511 s & 0.1737 s & 14.9\% \\
\bottomrule
\end{tabular}
}
\end{sc}
\end{small}
\label{tab:inference_time}
\end{center}
\vskip -0.15in 
\end{table}

\section{Ablation Study}\label{Abl}
This section evaluates the influence of the proposed framework's components and hyper-parameters on feature extraction and detection performance.

\begin{figure*}[t]
\begin{center}
\centerline{\includegraphics[width=\textwidth]{ 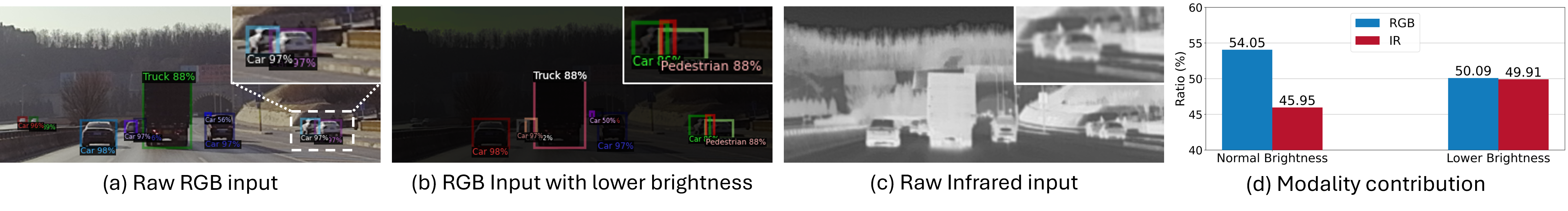}}
\vskip -0.15in
\caption{Impact of RGB brightness reduction in RGB-IR fusion during inference. (a)–(b): Normal and low-brightness RGB inputs. (c): Corresponding IR image. (d): Attention rates ($P_2^{\text{RGB}}, P_2^{\text{IR}}$), derived from $M_c^2$~\eqref{Eq5}, indicate the relative attention allocated to RGB and IR features, based on their ordering in the concatenated input feature vector. Trained on normal-brightness RGB, the model dynamically shifts focus toward IR under low-light conditions, enabling pedestrian detection in (b)–(c) that is missed in (a)–(c). Results are consistent across all $F_{P_i}$ levels, highlighting $M_c^k$ as an effective tool for interpreting multi-modal contributions.}
\label{dyn_ratio}
\end{center}
\vskip -0.2in
\end{figure*}

\subsection{Impact of Framework Components on Model Performance}
Table~\ref{tab:feature_contribution} summarizes the incremental contribution of the main components of the proposed approach, compared with the RGB-only DiffusionDet baseline reference. The components are progressively introduced according to the order in which they appear in the architecture (see Figure~\ref{arch_fig}).

The standalone integration of DCR-CBAM not only performs effective feature fusion leading to improved detection performance over the RGB-only baseline, but also enables monitoring of the attention allocated to features from different modalities, as shown in Figure~\ref{dyn_ratio}. The additional channel regularization term, $\theta_{M_c}$, yields further performance gains by emphasizing informative channels, which enhances the effectiveness of the channel reduction process.

\begin{figure}[t]
\begin{center}
\centerline{\includegraphics[width=\columnwidth]{ 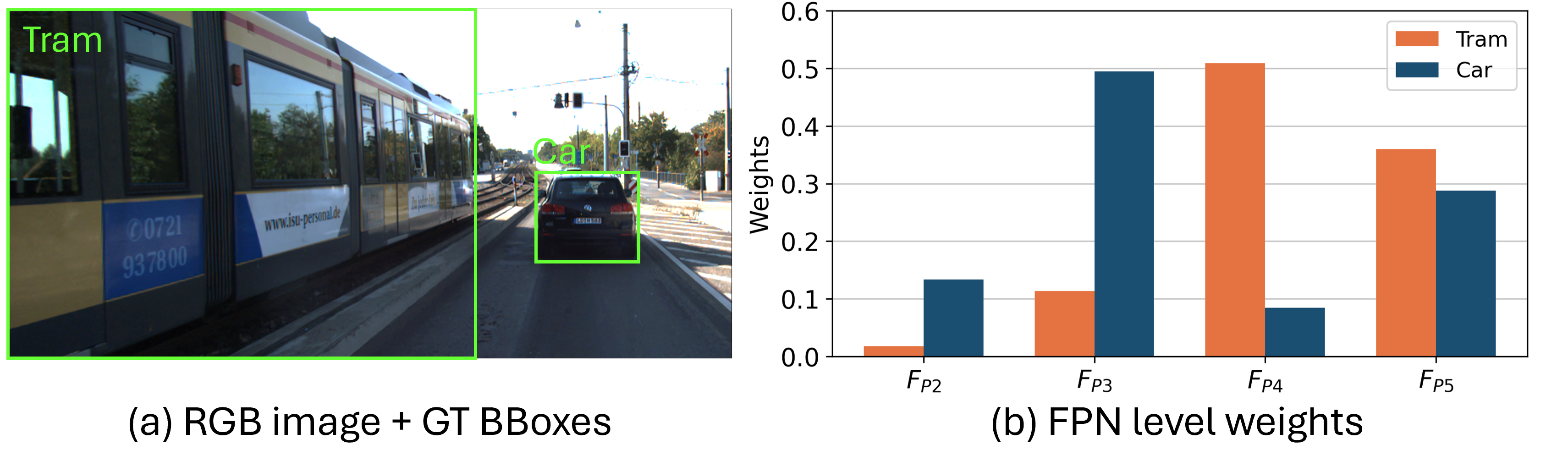}}
\vskip -0.15in
\caption{The dynamic behavior of DMLAB at inference mode. (a) shows object detections with corresponding class labels, while (b) plot illustrates the distribution of feature selection weights across different FPN levels. Unlike a deterministic FPN level selection approach, which would assign $F_{P2}$ to the car and $F_{P4}$ to the tram, DMLAB dynamically adjusts its weighting strategy based on feature importance.}
\label{dmlab_effect}
\end{center}
\vspace{-1.5em}
\end{figure}

Spatial regularization, $\theta_{M_s}$, enables focus on the most salient spatial regions based on ground-truth BBoxes. This refinement drives the feature map representation toward object detection by emphasizing informative areas. Combined with DMLAB, the model achieves the best detection probability of $69.2\%$. The weighted summation of multi-scale features by DMLAB further enriches the model's representation, delivering a more comprehensive feature map for the detection decoder (Figure~\ref{dmlab_effect}).

\begin{table}[t]
\vskip -0.14in 
\caption{Performance impact of each proposed component is evaluated by comparing the RGB-only DiffusionDet baseline with RGBX-DiffusionDet using RGB-D input.}
\label{tab:feature_contribution}
\begin{center}
\begin{small}
\begin{sc}
\begin{tabular}{l@{\hskip 1pt}c@{\hskip 1pt}c@{\hskip 1pt}c@{\hskip 1pt}c@{\hskip 1pt}l@{\hskip 1pt}}
\toprule
\multirow{2}{*}{Method} & \scriptsize{DCR-} & \multicolumn{1}{c}{\multirow{2}{*}{$\theta_{M_c}$}} & \multicolumn{1}{c}{\multirow{2}{*}{$\theta_{M_s}$}} & \multicolumn{1}{c}{\multirow{2}{*}{\scriptsize{DMLAB}}} & \multicolumn{1}{c}{\multirow{2}{*}{AP$_{50:95}$}} \\
& \scriptsize{CBAM}  &  &  &  &  \\
\midrule
\scriptsize{DiffusionDet} & & & & & 67.0 \\ 
\midrule
\multirow{4}{*}{\scriptsize{RGBX-DiffusionDet}} & \checkmark & & & & 67.7\posacc{0.7} \\ \cline{2-6}
& \checkmark & \checkmark & & & 68.0\posacc{1.0} \\ \cline{2-6}
& \checkmark & \checkmark & \checkmark & & 68.4\posacc{1.4} \\ \cline{2-6}
& \checkmark & \checkmark & \checkmark & \checkmark & \textbf{69.2}\posacc{2.2} \\
\bottomrule
\end{tabular}
\end{sc}
\end{small}
\end{center}
\vskip -0.14in 
\end{table}

\subsection{Impact of Channel Saliency Regularization on Modal Contribution Ratios}
\begin{figure}[t]
\begin{center}
\hspace*{-0.05\textwidth} 
\centerline{\includegraphics[width=0.9\columnwidth]{ 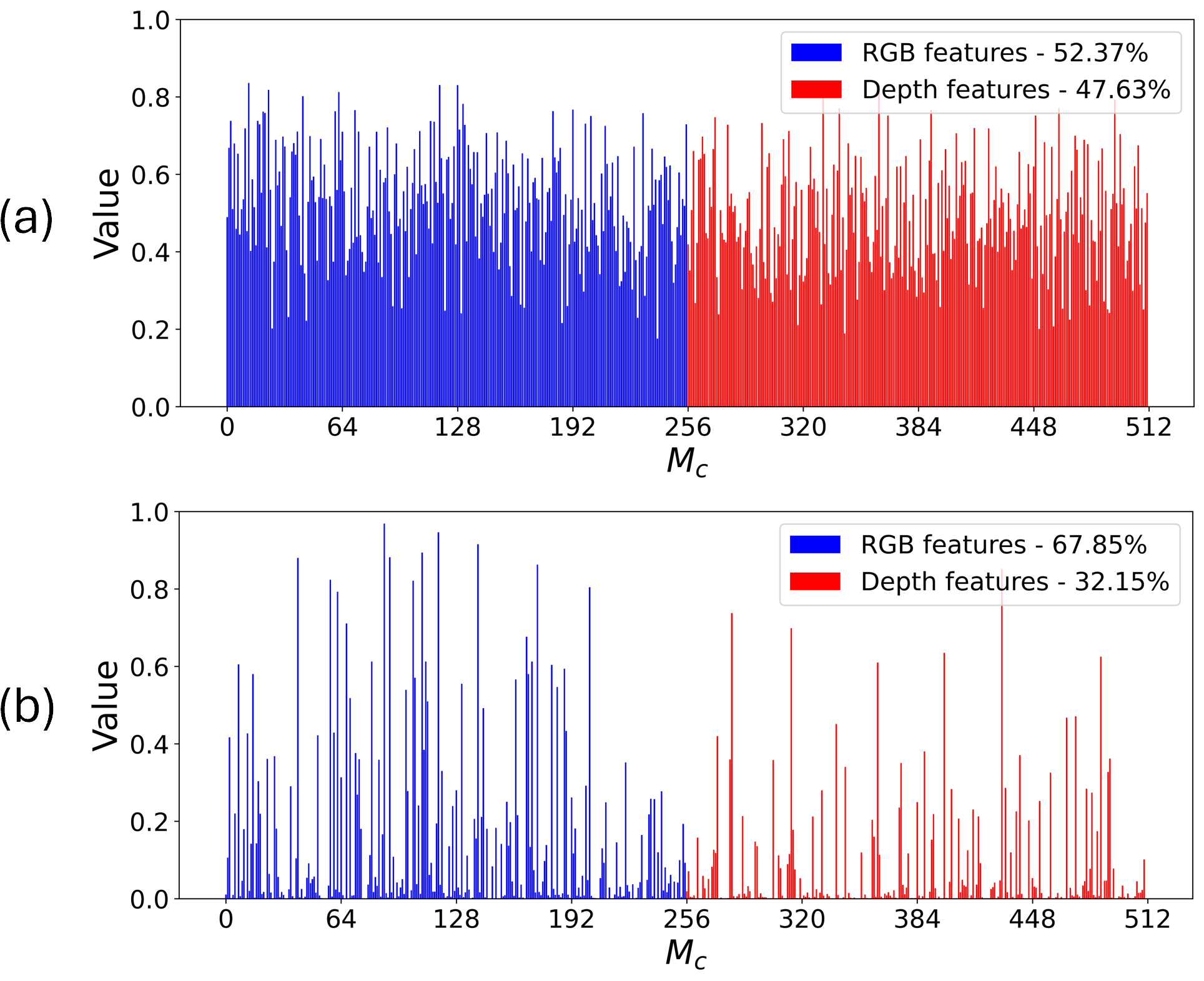}}
\vskip -0.1in
\caption{Effect of channel saliency regularization on $M^5_c$ maps. (a) Before regularization: RGB and depth contributions are nearly balanced. (b) After regularization: clearer modal separation emerges, with increased RGB dominance. The shift reflects reduced redundancy and improved fusion of complementary features.}
\label{creg_impact}
\end{center}
\end{figure}

The proposed approach not only lowers the energy of $M_C$, thereby promoting a richer feature representation, but also enhances the clarity of modal contributions. As illustrated in Figure~\ref{creg_impact}, the refined feature distribution effectively delineates the respective influences of RGB and depth modalities, mitigating redundancy and preserving the most salient features from each domain. This structured fusion process enables the model to exploit complementary information more effectively while maintaining architectural compactness. Consequently, the improved modal interpretability fosters more robust feature integration, ultimately enhancing object detection performance in multi-modal scenarios.

\subsection{Impact of Object-Focused Spatial Regularization on Detection Performance} \label{subsec:sp_regul}
\begin{figure}[t]
\begin{center}
\centerline{\includegraphics[width=0.95\columnwidth]{ 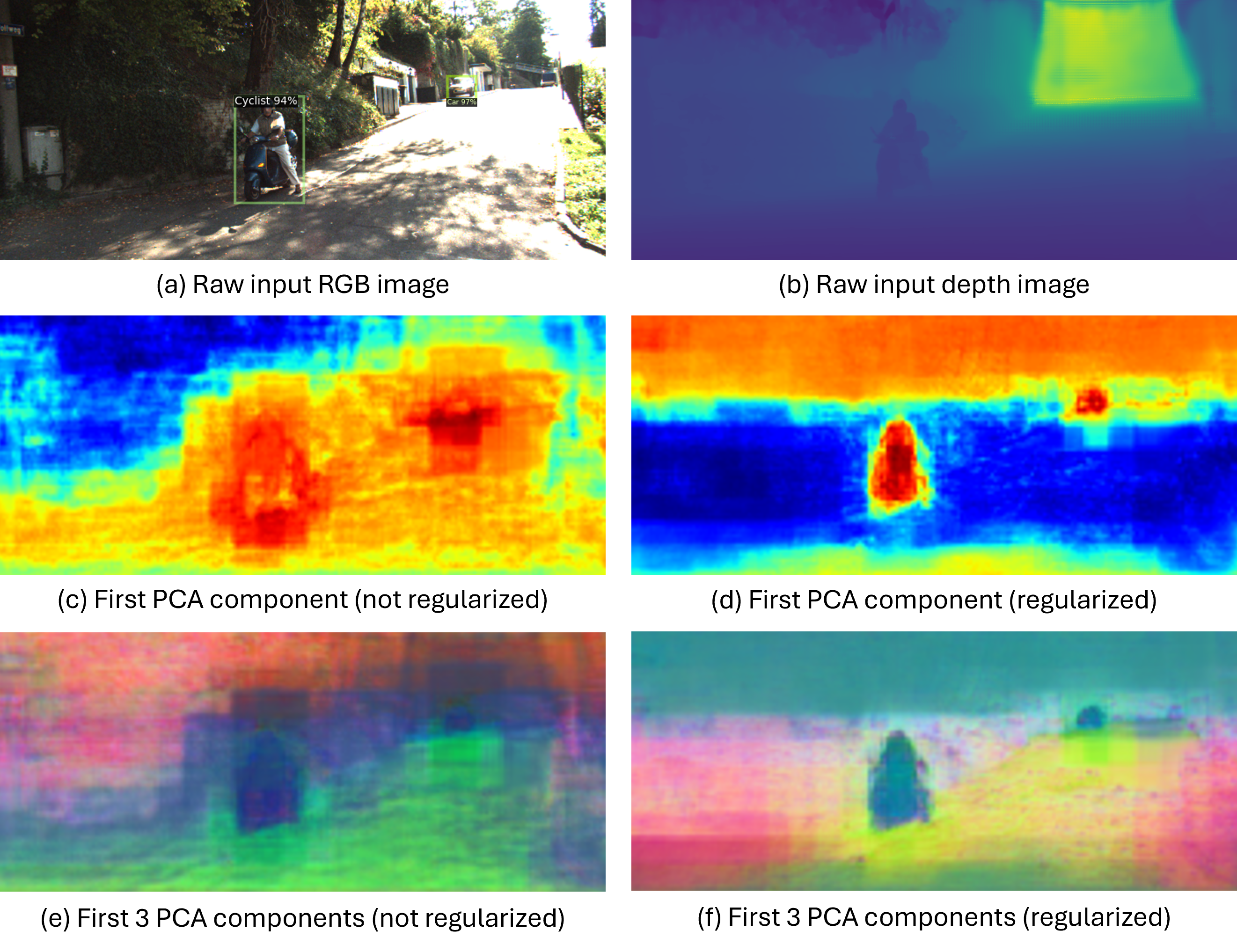}}
\vskip -0.1in
\caption{Effect of object-focused spatial regularization on $F_{P_2}$ features in an RGB-D setup. (a)–(b): input raw RGB and depth. (c)–(d): first PCA component of features without and with regularization. (e)–(f): first three PCA components. Regularization enhances focus on object-relevant regions and improves feature compactness.}
\label{FP2_pca}
\end{center}
\end{figure}

Figure~\ref{FP2_pca} shows the influence of object-focused spatial regularization on the feature representation in an RGB-D case. Without regularization (subplots c and e), extracted features typically capture global image characteristics. With regularization (subplots d and f), the model shifts focus toward object-relevant regions, enhancing task-specific feature extraction for detection. The influence of this modification is further evaluated using the principal component analysis (PCA). Comparing subplots (c) and (d), notice that the regularization increases the first PCA feature component energy from $36.1\%$ to $43.6\%$, and the energy of the first three PCA components from $59.9\%$ in (e) to $63.8\%$ in (f). These improvements indicate more compact and informative representations for object detection and suggest potential benefits for network compression~\cite{ma2019dimension}.

\begin{figure}[t]
\vspace{-1.5em}
\begin{center}
\centerline{\includegraphics[width=\columnwidth]{ 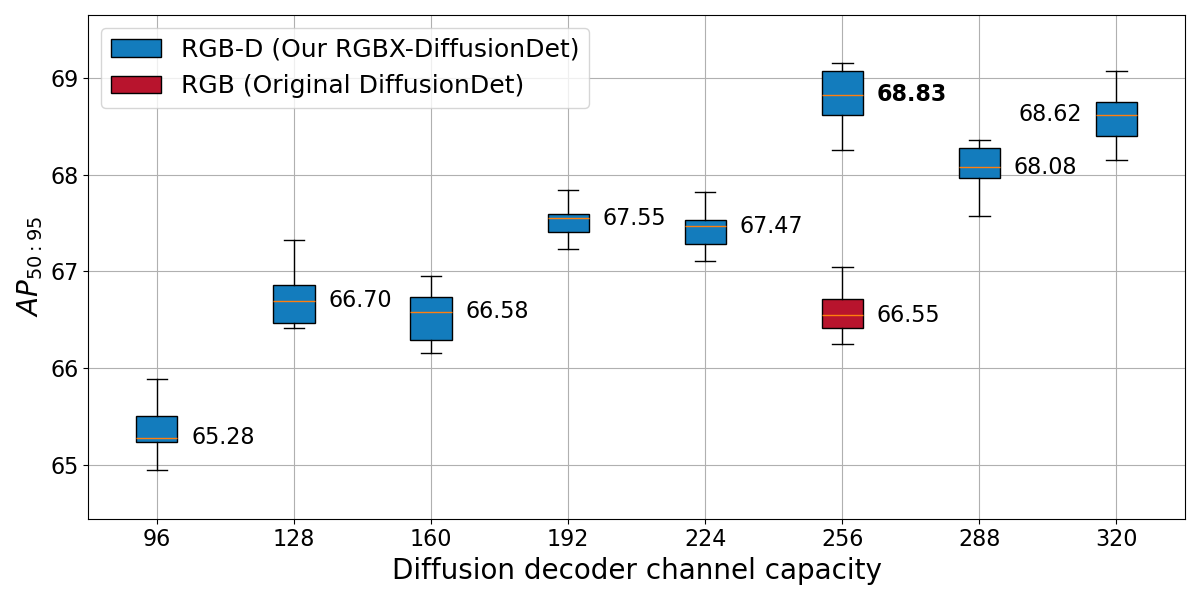}}
\vskip -0.1in
\caption{Influence of DCR-CBAM dimensionality reduction in RGB-D case on detection performance, compared with the RGB only case, marked in red, evaluated over $10$ independent simulations with random seeds.}
\label{box_fig}
\end{center}
\vskip -0.1in
\end{figure}

\subsection{Impact of DCR-CBAM Dimensionality Reduction}
Figure~\ref{box_fig} shows the impact of the DCR-CBAM dimensionality reduction module on model performance in the RGB-D setup. This experiment evaluates the effect of varying the number of channels $C'$ (Section~\ref{sec:DCR-CBAM}), on the performance of the proposed approach. Notice the performance degradation when using fewer channels, $C' = 96$, due to information loss, and when using more channels, $C' > 256$, due to increased capacity and the resulting model overfitting. Reducing the channel number to $224$ or $192$ outperforms the RGB-only DiffusionDet baseline, which uses $256$ channels. The optimal performance is achieved when the number of channels is $C'=256$. This experiment shows that adjusting the channel number efficiently mitigates overfitting and enhances generalization. Notice that the proposed DCR-CBAM, incorporating dimensionality reduction, maintains the decoder capacity equal to the original DiffusionDet for RGB data only.

\subsection{Corrupted RGB Camera} \label{Corrupted}
\begin{figure*}[t]
\centering
\includegraphics[width=\textwidth]{ 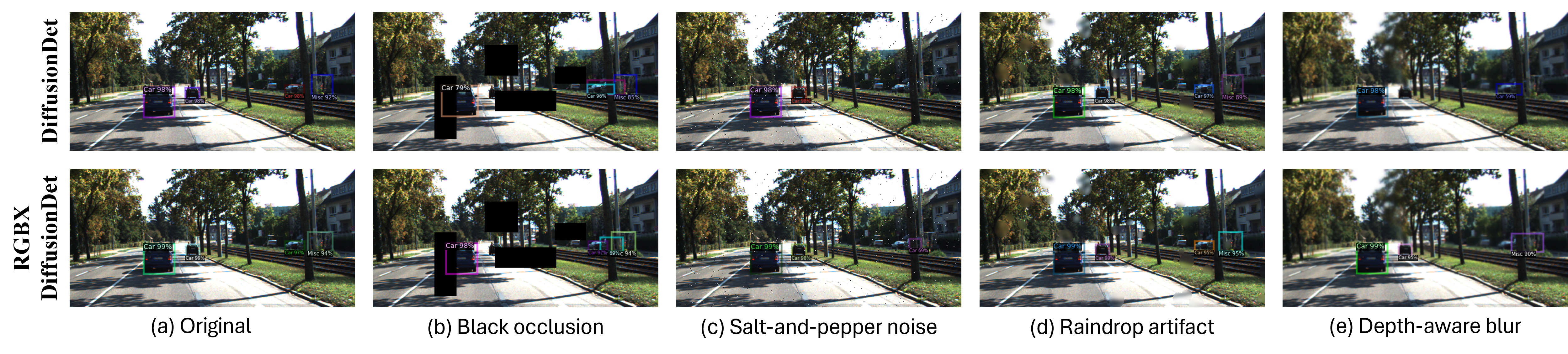}
\vspace{-0.5em}
\caption{Each column shows the same scene under different types of RGB corruptions. The top row presents the detection results from the baseline DiffusionDet model using only RGB, while the bottom row shows the results from our RGBX-DiffusionDet model. Despite corruptions in the RGB modality, the fused RGB-D input allows RGBX-DiffusionDet to maintain robust object detection performance with higher confidence scores and more accurate bounding boxes.}
\label{rgb_corruption_types}
\end{figure*}

To evaluate the robustness of RGBX-DiffusionDet under sensor degradation, we apply synthetic corruptions to the RGB input while keeping auxiliary modalities (e.g., depth) intact. This simulates real-world conditions where RGB sensors may fail due to environmental or hardware issues, while other sensors remain functional. We consider four corruption types (Figure~\ref{rgb_corruption_types}): \textit{black occlusion} (random black patches), \textit{salt-and-pepper noise} (bit-level interference), \textit{raindrop artifacts} (partial lens obstruction), and \textit{depth-aware blur}~\cite{huhle2009realistic} (depth-guided Gaussian blur). As shown in Table~\ref{rgb_corruption}, the RGB-only DiffusionDet baseline suffers a performance drop across all corruption types. In contrast, RGBX-DiffusionDet maintains strong detection accuracy, highlighting the model’s ability to exploit complementary modalities when RGB input is corrupted.

\begin{table}[t]
\begin{center}
\caption{Performance comparison in the RGB-D setting under simulated RGB corruptions at inference. 
}
\vspace{0.5em}
\begin{sc}
\resizebox{\columnwidth}{!}{%
\begin{tabular}{l | l | cc}
    \toprule
    Corruption type & [AP] & DiffusionDet & RGBX-DiffusionDet \\
    \midrule
    \multirow{2}{*}{Black occlusion}  & AP$_{50:95}$ & 48.6 & 56.3\posacc{7.7} \\
                                      & AP$_{50}$    & 70.2 & 77.2\posacc{7.0} \\
    \midrule
    \multirow{2}{*}{Salt-and-pepper noise} & AP$_{50:95}$ & 43.9 & 52.7\posacc{8.8} \\
                                           & AP$_{50}$    & 66.0 & 75.8\posacc{9.8} \\
    \midrule
    \multirow{2}{*}{Raindrop artifacts} & AP$_{50:95}$ & 61.0 & 66.1\posacc{5.1} \\
                                        & AP$_{50}$    & 85.6 & 88.7\posacc{3.1} \\
    \midrule
    \multirow{2}{*}{Depth-aware blur} & AP$_{50:95}$ & 51.3 & 58.5\posacc{7.2} \\
                                      & AP$_{50}$    & 74.5 & 81.1\posacc{6.6} \\
    \bottomrule
\end{tabular}
}
\end{sc}
\label{rgb_corruption}
\end{center}
\vspace{-0.5em}
\end{table}

\subsection{Diffusion Iteration}\label{subsec: dif_iters}
This subsection evaluates the efficiency of the diffusion process in RGBX-DiffusionDet as the number of iterations increases for all the datasets. Table~\ref{tab:step_iteration} shows that additional iterations progressively enhance detection performance. This validates the diffusion property, where each subsequent step further refines object predictions, regardless of the input modality or evaluation metric.

\begin{table}[t]
\begin{center}
\caption{Detection performance of the proposed RGBX-DiffusionDet as a function of the diffusion steps in RGB-D, RGB-P, and RGB-IR datasets. The detection performance was evaluated over $5$ simulations with different random seed numbers.}
\vspace{0.5em}
\begin{sc}
\resizebox{\columnwidth}{!}{%
\begin{tabular}{l | l | ccc}
    \toprule
    Dataset & [AP] & Step 1 & Step 2 & Step 3 \\
    \midrule
    \multirow{2}{*}{RGB-D} & AP$_{50:95}$ & 68.96 & 69.02\posacc{0.08} & 69.07\posacc{0.11} \\
                           & AP$_{50}$    & 90.50 & 90.75\posacc{0.25} & 91.06\posacc{0.56} \\
    \midrule
    \multirow{2}{*}{RGB-P} & AP$_{50:95}$ & 54.03 & 53.96\negacc{0.07} & 54.23\posacc{0.2} \\
                           & AP$_{50}$    & 71.59 & 71.63\posacc{0.04} & 71.77\posacc{0.18} \\
    \midrule
    \multirow{2}{*}{RGB-IR} & AP$_{50:95}$ & 58.06 & 58.13\posacc{0.07} & 58.29\posacc{0.23} \\
                            & AP$_{50}$    & 88.78 & 88.95\posacc{0.17} & 89.61\posacc{0.66} \\
    \bottomrule
\end{tabular}
}
\end{sc}
\label{tab:step_iteration}
\end{center}
\end{table}

\section{Conclusions}\label{Concl}
This work presents RGBX-DiffusionDet, an enhanced DiffusionDet framework that integrates auxiliary 2D modalities with RGB data via dynamic fusion modules and novel regularization terms. These components improve cross-modal representation learning while preserving the original decoder architecture and runtime efficiency.
Experiments across diverse RGB-X datasets demonstrate robust performance gains, making the approach applicable beyond autonomous driving to domains such as robotics, surveillance, and industrial automation.
Additionally, we extend a pixel-aligned RGB-P dataset with manual bounding box annotations, enabling its first use in object detection tasks. Despite added encoder complexity, inference remains lightweight, supporting the practicality of RGB-X fusion in real-world scenarios.

\section{Acknowledgments}
We thank General Motors for providing the RGB-Polarimetric (RGB-P) dataset~\cite{Baltaxe_2023_BMVC}, which enabled our research on RGB-P fusion. Our added bounding box annotations expand its utility for object detection and multi-modal learning.


\newpage
\bibliography{main}
\bibliographystyle{unsrt} 

\newpage
\appendix
\onecolumn

\section{Implementation Details}\label{ImplAondx}

\subsection{Model Architecture and Training}\label{ImplArchApndx}
The model utilizes ResNet$50$ backbones for both RGB and X feature extraction, initialized with pre-trained ImageNet-1K weights. The pretraining strategy enables us to train detection models that are robust in both small and large-scale data scenarios~\cite{jain2023}. Feature fusion integrates $256$ features from each backbone using the proposed DCR-CBAM, producing a $256$-channel representation for the diffusion decoder. The training was conducted for $80,000$ iterations on an NVIDIA A6000 GPU with a mini-batch size of $4$, using the AdamW optimizer for weight updates. Data augmentation was limited to random horizontal flips, and the number of proposal BBoxes was fixed at $N = 500$. Loss weighting parameters were set as $\lambda_{\text{label}} = 2.0$, $\lambda_{\text{bbox}} = 5.0$, and $\lambda_{\text{giou}} = 2.0$, while DCR-CBAM regularization was applied with $\lambda_{M_c} = 10^{-4}$ and $\lambda_{M_s} = 10^{-4}$. Model performance was evaluated using AP following the conventional object detection evaluation protocol \cite{lin2014microsoft}.

\subsection{Datasets} \label{ImplDataAondx}
The model was trained and evaluated using three datasets, with different sensor modalities:  
\begin{itemize}
    \item \textbf{RGB-D~\cite{geiger2012we}:} The KITTI dataset, split into $6,733$ training and $748$ test images, provided RGB and depth data in diverse road scenarios. The Viridis colormap from the Matplotlib library, commonly used in depth-related works for improved depth perception, was applied to the depth modality.  

    \item \textbf{RGB-P~\cite{Baltaxe_2023_BMVC}:} A total of $11,200$ training and $1,427$ test images were utilized, leveraging polarization cues to enhance object detection in dense urban environments. As described in~\ref{PolarEncodingApnx}, the polarimetric images were mapped to an RGB-compatible format, enabling seamless integration with the backbone architectures.

    \item \textbf{RGB-IR~\cite{Liu_2022_CVPR}:} The M3FD benchmark dataset, consisting of $2,940$ training and $1,260$ test images, incorporated both infrared and RGB modalities, improving robustness under varying lighting conditions. The infrared images in the M3FD dataset are provided in a single-channel $24$-bit grayscale format and are replicated across three channels to ensure compatibility.
\end{itemize}

\section{Polarimetric Data as RGB Image Representation}\label{PolarApnx}
Polarimetric imaging is a technique that captures information about the polarization state of light, providing unique insights into material properties, surface textures, and reflection characteristics~\cite{Baltaxe_2023_BMVC, blin2020new}. Unlike conventional RGB sensors that capture intensity across three color channels, polarimetric sensors detect the orientation and degree of polarization of light waves reflected from surfaces. This additional data can reveal subtle distinctions in surface composition and structure, such as differentiating between shiny and matte surfaces or detecting specific textures and materials that may not be easily identifiable in conventional RGB images. In the context of autonomous driving, where accurate perception of the environment is essential, polarimetric data can enhance object detection and scene understanding by providing richer visual information, especially in challenging lighting or reflective conditions.

\subsection{Polarimetric Sensor Functionality}

Polarimetric sensors capture light polarization by using polarizers oriented at specific angles (commonly $0$°, $45$°, $90$°, and $135$°) to obtain intensity measurements that vary according to the reflected light angle. These measurements are used to derive two polarimetric parameters:
\begin{itemize}
    \item \textbf{Angle of linear polarization (AoLP)}: representing the orientation angle of the polarized light wave, indicating the direction of polarization.
    \item \textbf{Degree of linear polarization (DoLP)}: describing the proportion of light that is polarized linearly, with values ranging from $0$ (unpolarized) to $1$ (fully polarized).
\end{itemize}

These parameters can be derived mathematically from the intensity measurements collected through the polarized filters. Considering the measured intensities at the four polarization angles, \( P_0 \), \( P_{45} \), \( P_{90} \), and \( P_{135} \), the overall intensity \( I \), AoLP, and DoLP are calculated as follows:

\begin{equation}
    I = \frac{P_0 + P_{45} + P_{90} + P_{135}}{2}\;,
\end{equation}

\begin{equation}
    \text{DoLP} = \frac{\sqrt{(P_0 - P_{90})^2 + (P_{45} - P_{135})^2}}{I}\;,
\end{equation}

\begin{equation}
    \text{AoLP} = \frac{1}{2} \arctan\left(\frac{P_{45} - P_{135}}{P_0 - P_{90}}\right)\;.
\end{equation}

This formulation allows polarimetric data to be quantified in terms of AoLP and DoLP values, which offer a unique set of visual information complementary to RGB data.

\subsection{Encoding Polarimetric Data in an RGB-Compatible Format} \label{PolarEncodingApnx}

This work represents polarimetric data within conventional image-processing architectures in an RGB-compatible format. This is achieved by encoding the AoLP and DoLP values into a three-channel representation:

\begin{equation}
    P = \left[ \sin(2 \cdot \text{AoLP}), \cos(2 \cdot \text{AoLP}), 2 \cdot \text{DoLP} - 1 \right]\;.
\end{equation}

\begin{figure*}[hb]
\begin{center}
\centerline{\includegraphics[width=0.6\textwidth]{ 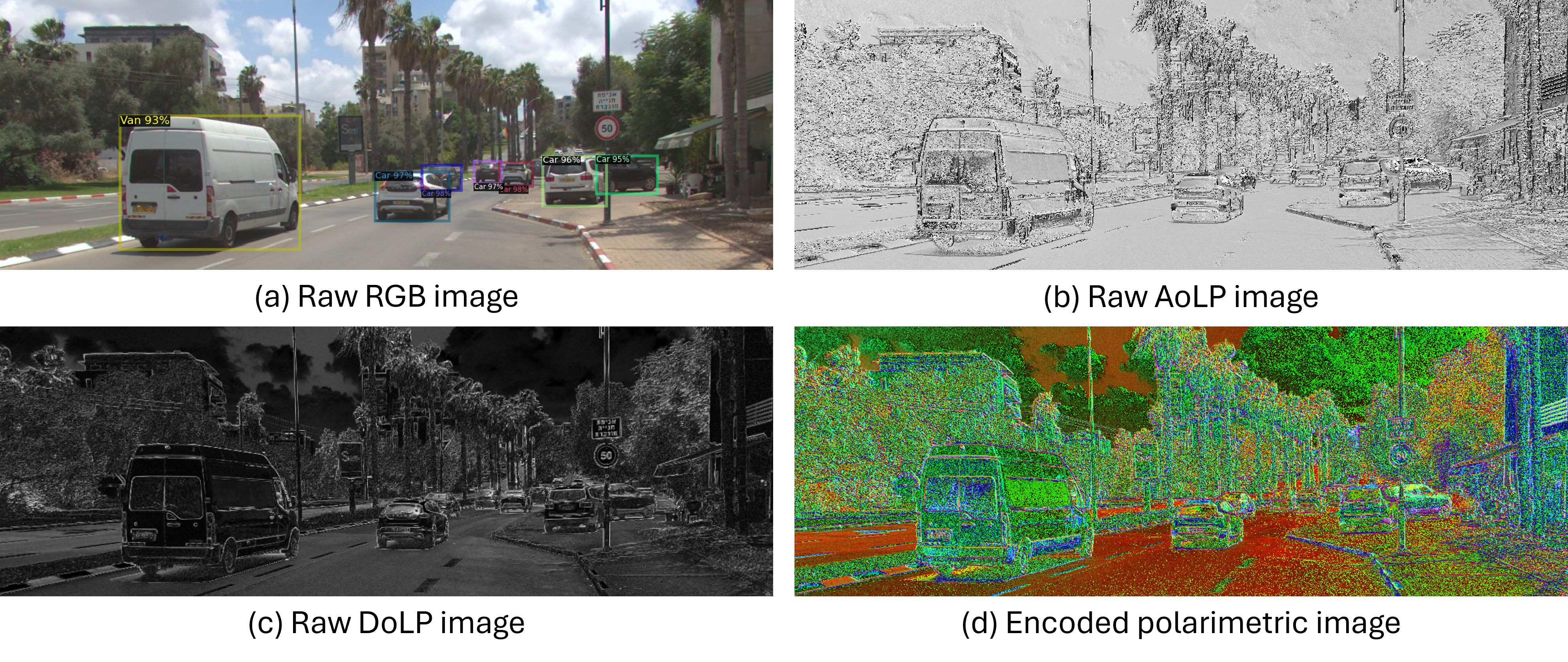}}
\caption{
RGB-P object detection within our proposed framework. (a) shows the RGB image with detected objects and their bounding boxes. (b) and (c) depict the AoLP and DoLP components, respectively. These polarimetric features are encoded into a composite representation in (d). The encoded image (d) and the RGB image (a) serve as inputs to the proposed framework.}
\label{polarExmple}
\end{center}
\end{figure*}

This transformation effectively maps the cyclic and continuous nature of AoLP and DoLP into a format that can be processed as three image channels (Figure~\ref{polarExmple}). Key aspects of this encoding are:

\begin{itemize}
    \item \textbf{Handling AoLP’s cyclic nature}: By converting AoLP into sine and cosine components, \(\sin(2 \cdot \text{AoLP})\) and \(\cos(2 \cdot \text{AoLP})\), we capture the cyclic quality of the polarization angle, ensuring that the model interprets AoLP smoothly without abrupt changes due to its periodicity.
  
    \item \textbf{Normalizing DoLP for Compatibility}: The DoLP parameter is scaled to \([-1, 1]\) as \(2 \cdot \text{DoLP} - 1\), aligning with the common data ranges expected by RGB-trained models.

    \item \textbf{Three-Channel Representation}: This format effectively integrates the unique polarization data as an RGB-equivalent structure, making it compatible with existing deep learning models designed for RGB data. 
\end{itemize}

By transforming polarimetric data into this format, we can leverage it in conventional vision tasks, allowing object detection and scene understanding models to benefit from additional material and texture information without architectural modifications. This encoding preserves the crucial details of AoLP and DoLP, maximizing polarimetric data's efficiency in applications that conventionally use RGB inputs only.

\end{document}